\documentclass{sig-alternate}
\usepackage{amsmath,amssymb,hyperref,graphicx,url,times,fleqn,flushend,times,subfigure}
\usepackage{epstopdf}
\usepackage{breqn}
\usepackage{syntonly}
\usepackage{color,soul} 
\usepackage{comment}
\newcommand{\OMIT}[1]{}

\hypersetup{%
   pdftitle={},
   pdfauthor={Beard},
   pdfkeywords={latex, template},
   bookmarksnumbered,
   pdfstartview={FitH},
   colorlinks=true,
   breaklinks=true,
   citecolor=blue,
   filecolor=blue,
}%

\begin{document}
\title{Deep Neural Net with Attention for Multi-channel Multi-touch Attribution}

 \author{
    Ning Li\\
    ningli30@uw.edu
  \and
    Sai Kumar Arava\\
    arakumar@adobe.com
  \and
    Chen Dong\\
    chedong@adobe.com
    \and
    Zhenyu Yan\\
    wyan@adobe.com
  \and
    Abhishek Pani\\
    apani@adobe.com
}

\maketitle

\begin{abstract}
Customers are usually exposed to online digital advertisement channels, such as email marketing, display advertising, paid search engine marketing, along their way to purchase or subscribe products(aka. conversion). The marketers track all the customer journey data and try to measure the effectiveness of each advertising channel. The inference about the influence of each channel plays an important role in  budget allocation and inventory pricing decisions.
Several simplistic rule-based strategies and data-driven algorithmic strategies have been widely used in marketing field, but they do not address the issues, such as channel interaction, time dependency, user characteristics.
In this paper, we propose a novel attribution algorithm based on deep learning to assess the impact of each advertising channel. We present Deep Neural Net With Attention multi-touch attribution model (DNAMTA) model in a supervised learning fashion of predicting if a series of events leads to conversion, and it leads us to have a deep understanding of the dynamic interaction effects between media channels. DNAMTA also incorporates user-context information, such as user demographics and behavior, as control variables to reduce the estimation biases of media effects. 
We used computational experiment of large real world marketing dataset to demonstrate that our proposed model is superior to existing methods in both conversion prediction and media channel influence evaluation.   

\end{abstract}



\section{Introduction}
Online advertising has grown exponentially over the past few years due to the wide spread usage of internet across the world.
The marketers track customer journeys as they are exposed to different online media channels(e.g. email, display, paid search) before they make the conversion at the end. 
Companies allocate marketing budgets to promote their business through these multiple online campaigns among different channels. To get maximum return on investment on the spend of online ads, marketers have to optimize their budget allocation among different media channels based on their value. How to measure the value of ads spend, however, is not trivial for marketers. The problem of measuring the influence of each campaign or channel on a conversion is referred as attribution problem \cite{kannan2013Experiment} . 
\begin{figure}
\centering
\includegraphics[width=.8\linewidth]{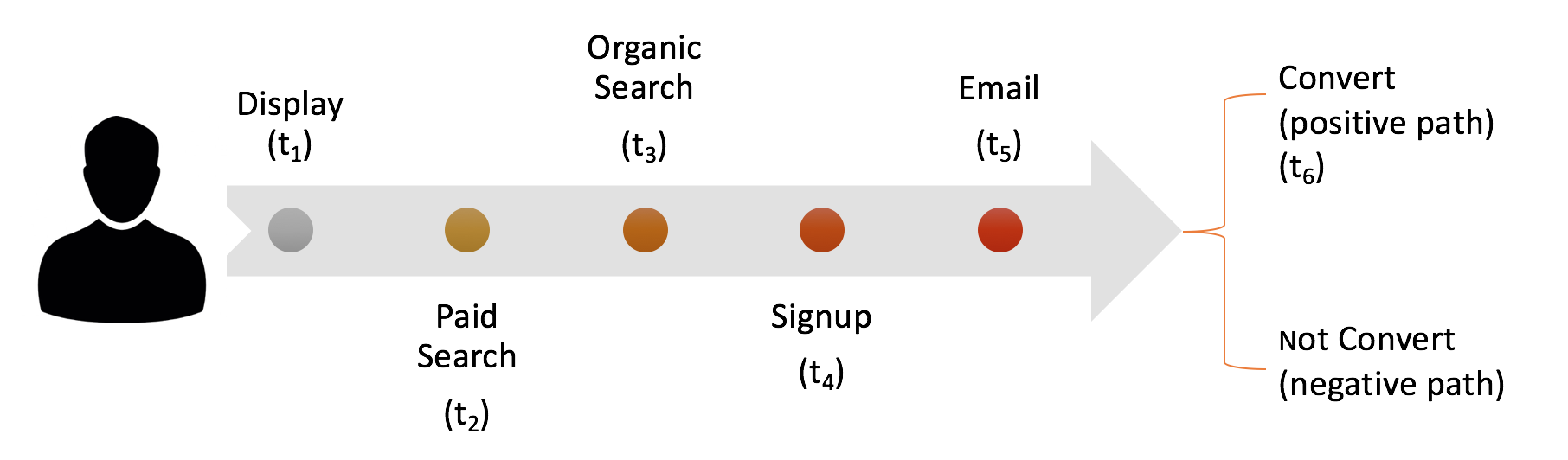}
\caption{A possible behavioral customer journey in an online advertising system. Here, the user is exposed to display, paid search and email touch points, but he or she may choose to make conversion or not at the end }
\label{fig:human_conversion_path}
\end{figure}

As shown in Figure[\ref{fig:human_conversion_path}], a user may be exposed to email, display, paid search ads before the users converts. Each ad has a relation with the user's final conversion decision. In such a case, the marketer faces a dilemma of assessing the contribution of each channel to user's conversion.

Marketers have applied simple rule-based heuristics to solve attribution problem in the past. First or last touch point approach ignore the effects of other channels; equal weight approach assume equal contribution from each channel, which ignores the channel difference; time-decayed attribution algorithm assumes that the credit decays based on a decay parameter which is simply based on intuition without data support. 

In order to rectify the above pitfalls, data-driven attribution models have been introduced in recent years. 
In this paper, we propose a data-driven multi-touch attribution and conversion prediction model denoted as deep neural net with attention for multi-touch attribution (DNAMTA) that outperforms the other approaches in terms of both conversion prediction and attribution analysis.



\section{Related Work}
In order to overcome the drawbacks of rule-based heuristics, data-driven algorithmic models were proposed. Shao et al. \cite{Shao2011KDD} propose a bagged logistic regression method and compares it with a probabilistic model. They predict conversion rate using count of ad occurrences and uses weights as credits for attribution analysis. While bagging provides stable estimates and better accuracy than probabilistic model, they do not have an interpretable model and ignores temporal factor. Dalessandro et al. \cite{brian2012causal} propose causal inference methods to achieve interpretability. They used additional marginal lift of each ad as credits. Since their method was computationally difficult, under some assumptions, they were unable to estimate causal parameters. 
Ji et al. \cite{Ji2016Probabilistic} adopt a probabilistic framework to remove the presentation biases. However, they do not directly measure the effect of ad exposure. 
Zhang et al. \cite{Zhang2014Survival} propose data-driven multi-touch attribution with survival theory but do not consider user characteristics.Ji et al. \cite{Ji2017Survival} use hazard rate to reflect the influence of an ad exposure. However, they assume that the impact of ad exposures is additive and fades with time. Abhishek et al. \cite{abhishek2012HMM} propose multistage model of consumer response to advertising activity that addresses the problem of temporal dynamics of ad exposure. 
However, their framework is difficult to achieve model scalability, besides, higher order markov chains are hard to be implemented for better model accuracy. 

Deep Learning \cite{Goodfellow-et-al-2016} have been used extensively in image \cite{larochelle2010learning}, speech recognition and language translation \cite{bahdanau2014neural} to achieve state-of-art results. Attention mechanism embedded with Neural Network has been successfully applied in vision and NLP field \cite{denil2012learning,bahdanau2014neural}, as attention mechanism can emphasize the important features along the time-series observations.   
These novel ideas, however, are not yet used to tackle problems like attribution.



\section{Notation and Problem Formulation}
\label{notation}
\par We formalize the attribution problem as follows. An event is either a conversion or a touchpoint. Each customer path consists of events from multiple advertising channels. Let $x_t$ denote the $t^{th}$ event the user is exposed to in the path and $x_t\in \mathbb{M}$, $\mathbb{M}$ is the set of all the touchpoints that we are interested in. Thus, a single customer sequence path $P_i$ can be represented as $P_i = \{x_1, x_2, \dots, x_T\}$, $T$ is the length of the sequence. $t$ represents the relative order of the event in the sequence, instead of the absolute event occurrence time. Beyond that, each event is also associated with some structure information, such as occurrence time, which can be formalized as another sequence $\{U_1, U_2, \dots, U_T\}$. In addition to these dynamic sequence information, some static information which is unlikely to be changed during the conversion journey, such as gender, age, sign-up date  etc., are represented as control variables $C_i$. A customer path will be treated as positive if it ends with conversion($Y_i =1$), otherwise it's a negative(non-conversion) path($Y_i = 0$). Assuming each touchpoint $x_t$ has attribution value $a_t$, then $\sum_{t=1}^Ta_t = 1$. The objective of this attribution problem is to estimate attribution value $a_t$ which represents the touchpoint $x_t$'s contribution towards a successful conversion.

\par  To make this problem more mathematically well-defined, we use probabilistic reasoning to explain customer's conversion decision, i.e. we want to find how likely a path will end up with conversion if it is exposed to a sequence of touchpoints  $P_i$ and its corresponding control variable  $C_i$. We denote this as conditional probability $P(Y_i|P_i, C_i)$. 
According to Bayes formula, $P(Y_i|P_i, C_i) = P(Y_i, P_i | C_i)/P(P_i|C_i)$ and in order to get a good inference of this conditional probability we should have a good estimate of two components: $P(Y_i, P_i |C_i)$ and $P(P_i|C_i)$. $P(Y_i, P_i|C_i)$ can be estimated by maximum likelihood estimation(MLE) from the data. Since $P_i$ is a dynamic sequence observation with varying length, estimating $P(P_i|C_i)$ is difficult. Furthermore, if we use a naive one-hot representation by aggregating through time, it ignores the time variance information. Therefore, it's necessary to have a better representation of $P_i$, that helps to estimate probability $P(P_i|C_i)$ and  $P(Y_i | P_i, C_i)$ easily. We use a learning function $f$ to approximate this conditional probability $P(Y_i|P_i, C_i) = f(C_i, \{x_t\}_{1:T})$. Thus the underlying structure for attribution of each touchpoint can be estimated from this learning function.  


\par Attribution problem is complex as hidden interactions between touchpoint needs to be modeled. Besides, contribution of touchpoint decreases with the increasing time lag(defined as the duration between the occurrence time and the end time) in a path. This typical time decay property is a common business assumption, which is unlikely to be captured by general linear model. Lastly, control variables like gender, age, sign up date etc. can also affect customer journey. 

\par We propose a general deep learning framework in order to solve the above three challenges: DNAMTA. This model has three advantages: 1) \textit{DNAMTA with attention} is a Long Short Term Memory (LSTM) based deep sequential model, which is well known for capturing the long time dependency of sequence observations\cite{hochreiter1997long}. Further, attention mechanism is used to capture the touchpoint contextual dependency.  2) Survival time-decay functions are introduced in \textit{DNAMTA with timedecay} to explicitly model the timedecay assumption. 3) \textit{DNAMTA fusion} model  can combine static information of user as control variables with dynamic touchpoint observations. 
\vspace*{-.2cm}
\section{ Deep Neural Network with Attention for Attribution}

\par
\begin{figure}
\includegraphics[width=.9\linewidth]{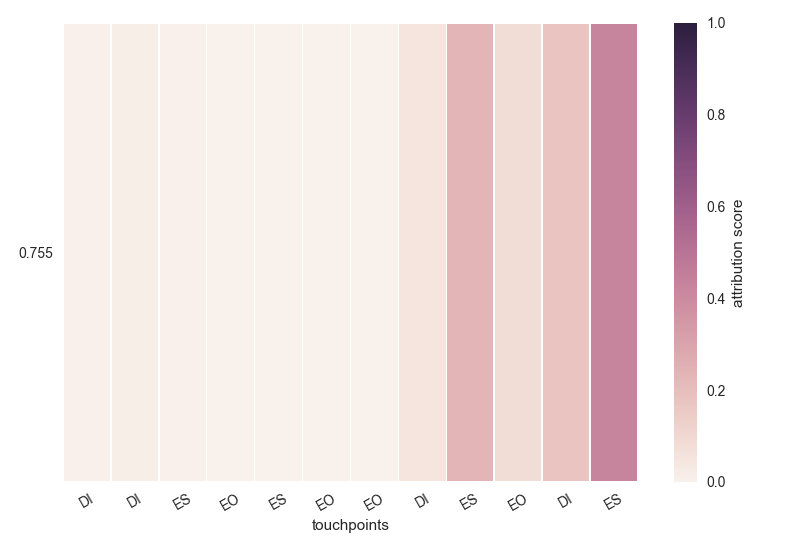}
\caption{A heatmap visualizing the contribution of each touchpoint in a specific customer conversion path. From left to right, customer journey starts from the first event to the end of conversion, all events are coded by:display_click(DC), display_impression(DI), email_click(EC), email_open(EO), email_sent(ES), paid_search(PS). Y-axis indicates the conversion probability from the prediction model. The darker the color for a touchpoint, the higher influence of the corresponding touchpoint is.  }
\label{fig:att_heatmap}
\end{figure}

 In a sequence of observations of touch points, same touchpoint
may be differentially important at different time locations and at
different frequency of occurrence. Our model introduces attention mechanism that
lets the model to pay more or less attention to individual
touchpoints when constructing the representation of the customer
path. To demonstrate the idea, let's take a look at Figure(\ref{fig:att_heatmap}), which is a positive path where the customer finally made a conversion at the end. This customer has been exposed to a sequence of advertising events before a conversion decision is made. Each touchpoint is allocated different contribution value according to our model. The contribution of touchpoint "Email Sent" varies at different observation locations. Besides, touchpoint "Email Sent" has totally different importance compared with other touchpoints, such as "Display Impression". Details of our proposed model will be covered in Section[\ref{model}]. 
\par Attention serves two benefits: it not only provides us
reasonable better performance, but also gives insight on how
touchpoint contributes to the conversion decision at any specific time
which is the most valuable part of an attribution conversion problem.  LSTM could help us handle capture the hidden underlying complex interaction patterns.


\vspace*{-.2cm}
 \section{Model}
 \label{model}
 \label{fig:att_model}

\vspace*{-.2cm}
\subsection{Touchpoint Input Layer}
\par The input for the model is a touchpoint sequence path $P$ with one-hot representation of touch points
$x_{t}, t\in [0, T], x_{t}\in \mathbb{R}^{v_{tp}}$, where $v_{tp}$ is the total number of all possible touchpoints we are interested in and $T$ is the length of the sequence, which varies for different sequences. Note that this sequence only considers absolute order, the real temporal difference between each touchpoint pair could be different. Detailed information about dealing with temporal relation will be discussed in later section [\ref{att_timedecay}]. 
\vspace*{-.2cm}
\subsection{Touchpoint Embedding Layer} 
\par Given a path $P$ in the above format, we first transform the one-hot representation of the
touchpoint at step $t$ to a dense vector through an embedding matrix $W_e\in \mathbb{R}^{v_e \times v_{tp}}$ by $e_{t} =
W_ex_{t}$. Specifically, $t^{th}$ column of embedding matrix $W_e$, which is a vector of $v_e$ dimension, is the continuous representation vector of step $t$ touchpoint observation. 
\par Traditional one-hot representation or bag-of-words like feature representation are simply counting statistics, which ignore touchpoint contextual similarities and suffer from sparsity in representation. Touchpoint embedding aims to quantify and categorize hidden contextual similarities between each touchpoint based on their distribution in large samples of touchpoint paths. 
\vspace*{-.15cm}
\subsection{Variable-depth LSTM Layer}
\par We use LSTM \cite{hochreiter1997long} to obtain another level of representation of touchpoints by using embedding layer output $\{e_{1}, \dots, e_{T}\}$, and therefore incorporate the
contextual information in the historical observations.  Each LSTM block updates current hidden state output $h_{t}\in \mathbb{R}^{v_h}$ based on embedding output $e_{t}$ and previous hidden state output $h_{ t-1}$ through the formula 
\begin{eqnarray}
h_{t}& = & \mathcal{H}(e_{t}, h_{t-1}), ~ t\in [0, T]
\label{eq:hidden_state}
\end{eqnarray}
\par In Formula(\ref{eq:hidden_state}), $\mathcal{H}$ is a nonlinear transformation function, which has various definitions according to practical problems.
\par Now each $h_t$ can be considered as a new representation of $t^{th}$ touchpoint by overviewing all historical touchpoint records, so conceptually $h_t$ is able to better describe the context meaning of touchpoint in the specific path compared with the raw embedding vector $e_t$, which is unaware of past information. This is important for customer conversion journey, since the order, frequency and long-term dependency of touchpoint exposure could have a high impact on their final conversion decision. 
\vspace*{-.2cm}
\subsection{Touchpoint Attention Layer}
 \par 
We introduce attention mechanism to find touchpoints that are
important to the conversion and aggregate the
representation of those informative touchpoints to form a path
vector.  Yang et al. \cite{yang2016hierarchical} proposed hierarchical attention mechanism for text sentiment analysis. We shall leverage this idea in our case. Specifically, 
\begin{eqnarray}
v_{t} & = & tanh(W_v h_{t} + b_v)\\
a_{t} & = & \frac{exp(v_{t}^Tu)}{\sum_{t}exp(v_{t}^Tu)} \label{eq:att}\\
s & = & \sum_t a_{t} h_{t}
\end{eqnarray}
We first feed the touchpoint representation $h_{t}$ through
a one-layer multilayer perceptron(MLP) to get $v_{t}$ as a hidden representation of $h_{t}$, then we measure the importance of the word
as the similarity of $v_{t}$ with touchpoint context vector $u$ and
get a normalized importance weight $a_{t}$ through a softmax
function. We can notice that by design $a_t > 0$. The advantage of this construction is that the contribution of every touchpoint is always positive. After that, we compute the path vector $s$ as the weighted
sum of the touchpoint representation based on the non-negative
weights. Actually, $s$ is the convex combination
of all $h_{t}$. The context vector $u$ can also be seen as a high level representation of a fixed sequence based on our domain knowledge about touchpoint
importance, campaign marketers can custom their attribution
model by constraining vector $u$. The context vector $u$ can either be fixed or be randomly initialized and jointly learned during the process. We use the latter approach in our modeling. 
\subsection{Touchpoint Path Classification}
\label{tp_classification}
\par  In our attribution conversion problem, some customer touchpoint journeys end up with conversions, these paths are treated as positive paths, otherwise, they are negative paths. With these labels, we can consider this attribution conversion learning problem as binary classification problem in the new path vector space. The path vector $s$ is a high
level representation of the customer touchpoint journey by combining hidden outputs and attention weights. 
\begin{eqnarray}
p& = & sigmoid(\sigma(W_c^T s)+ b_c)
\end{eqnarray}
where $W_c \in \mathbb{R}^{v_h}$ and $\sigma(\cdot)$ is nonlinear activation function ReLU $\sigma(x) = \max{(0, x)}$. In common binary classification problems, the probability for predicting the sequence observation path positive is usually the sigmoid function for linear combination of features. In attribution conversion problem, with some exposure of advertising channels, the probability for customers to make conversion decision is always greater than those without any exposure, which means the contribution of touchpoint for conversion is always nonnegative. Activation function ReLU is mathematically fit for this practical constraint.

\subsection{Time Decay Attention Layer}
\label{att_timedecay}
\par Attention mechanism is widely used in NLP problems where the distance between each word is relative, depending on the word counts between them. 
We should consider exact time gap in attribution problem, since the time gaps between each touchpoint vary a lot, from hours to even months. This difference of time gaps could affect the connection strength of nearby touchpoints and further impact the final conversion. 
Therefore, we introduce the time decay attention layer by combining time decay information, inspired by the idea in \cite{wooff2015TimeWeighted}. Basically each touchpoint sequence observation has its occurrence time, the time gap difference between the occurrence time and the end time defined as $T_t$. The smaller $T_t$ is, nearer is the occurrence time to end time. We assume the touchpoint contribution decreases when the occurrence time is far away from the end time. We penalize each attention weight described in component in Formula(\ref{eq:att}) by non-increasing timedecay function. Detailed formula can be referred as below:

\begin{eqnarray}
v_{t} & = & tanh(W_v h_{t} + b_v)\\
a_{t} & = & \frac{exp(v_{t}^Tu - \lambda T_t)}{\sum_{t}exp(v_{t}^Tu - \lambda T_t)} \label{eq:att_decay}\\
s & = & \sum_t a_{t} h_{t}
\end{eqnarray}

where $\lambda > 0$  is the decay parameter, it can be predefined based on data analysis of past customer conversion trend, or it can also be randomly initialized in model and directly learned from data. 

\subsection{Fusion Model}
\par As we have mentioned in previous section, attribution conversion models usually try to establish the relationship between advertising channels and final conversion. However, customer characteristic information such as gender, age and some other static information may affect the touchpoint exposure and the conversion engagement. \cite{rosenbaum1984reducing} points out that the confounding effects from these features could affect the distribution of conversion rate. For example, free signup is a promotion strategy from company to encourage customer to make conversion. Generally there are two situations when a conversion rate may peak: First, when  customers free signup and second,when this free signup trial expires. Therefore, it's necessary for us to take these control variables into consideration, which helps us to minimize the potential bias inference effects. 
\par
\begin{figure}
\centering
\includegraphics[width=1.0\linewidth]{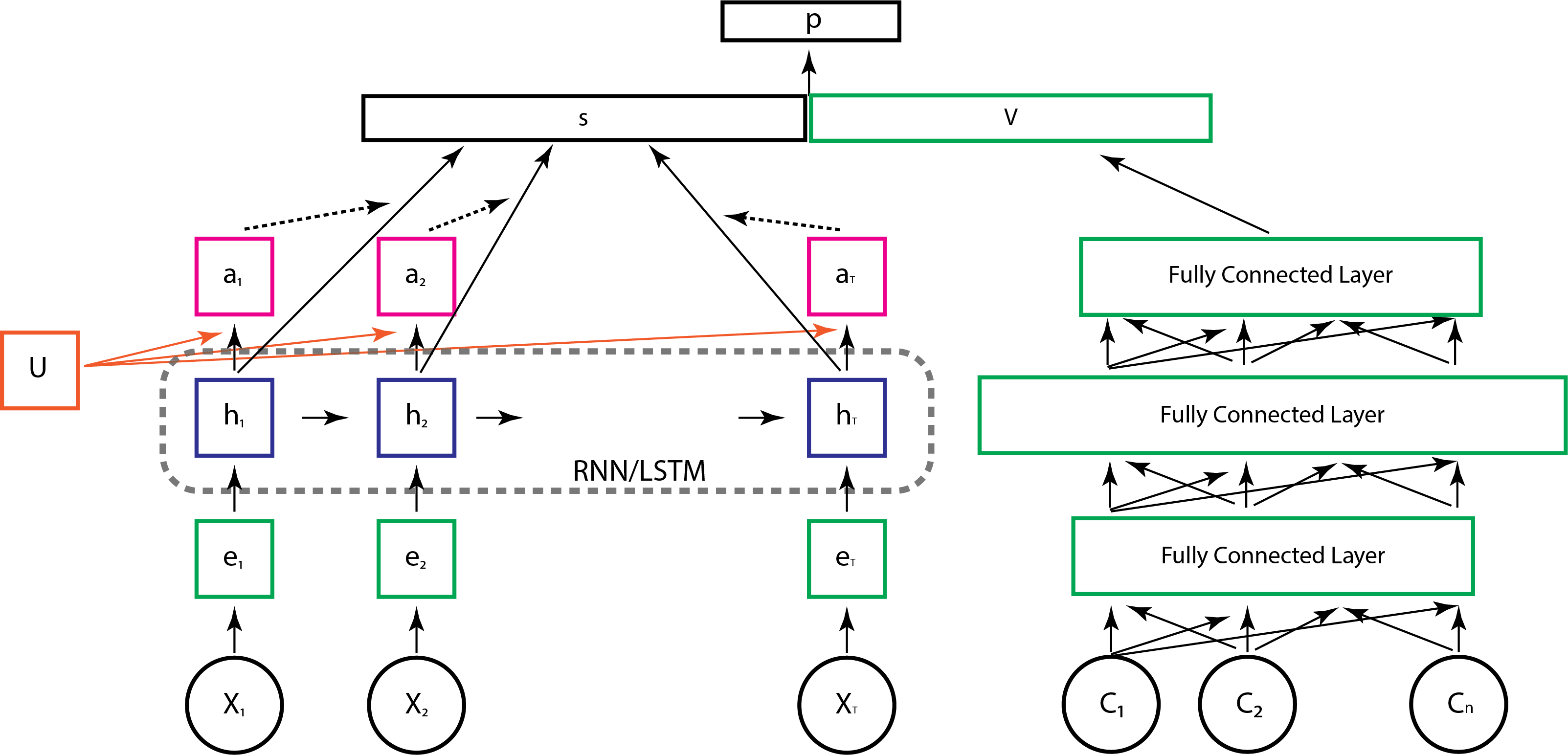}
\caption{The structure of DNAMTA fusion model, including three parts: sequence encoder, control variable encoder and sequence classification}
\label{fig:att_fusion}
\end{figure}

 However, the number of control variables in real attribution conversion problems can be very large, which increases the difficulty of the variable selection among these control sets. Besides, a simple linear add-on may not fit for describing the complex relationship between the factors and conversion. In order to account for these two problems, we propose a fusion model, which is built on the original DNAMTA model by introducing another deep neural network for control variable learning. In Figure(\ref{fig:att_fusion}), deep neural network modeling control variables is on the right hand. It aims to learn a sophisticated feature vector representation by going through several dense fully connected layers, which can capture the underlying structure. Later we concatenate the customer touchpoint path representation vector and the control variable vector before we apply it to classification layer. The touchpoint path classification formula will be changed
\begin{eqnarray}
p& = & sigmoid(\sigma_1(W_{c_{tp}}^T s)+ \sigma_2(W_{c_{ntp}}^T v) + b_c) \label{eq:fusion}
\end{eqnarray}
where $\sigma_1$ is still the RELU function as mentioned in Section [\ref{tp_classification}], and $\sigma_2$ is just identical function. 

\section{Experiments}
\par In this section, we present our experiments to compare different attribution models with DNAMTA. We also discuss the corresponding results and model interpretation for attribution. 
\subsection{Data}

\par We ran our experiments on large event data set of a marketing organization with three primary channels display, email and paid search. We have 6 different touchpoints: display click, display impression, email click, email sent, email open and paid search. It contains 426853 records with history of 57 days including conversion day. Each record represents a customer's journey, if this journey ends up with conversion action before the given data collection time, it is regarded as positive path; otherwise it's labeled as negative. Even though customer may convert in the future, this kind journey is still not positive based on our definition. Due to the heavy imbalanced distribution of positive and negative paths in real dataset, we down sampled the negative path records to get the dataset with balanced labels. Each path record is associated with a free-signup date, a sequence of dates for each touchpoint event and a sequence of frequency of occurrence. A visit duration window is applied to multiple visits from the same advertising channel: subsequent visits are ignored if they occur within a short time. 
\par We randomly split this data into two sets: 80\% for training and 20\% for testing. All experiment comparison results are based on the test dataset.

\subsection{Model Settings and Implementation}
As mentioned in \cite{Eva2014Graph}, we will mainly focus on predictive accuracy(AUC) and interpretability. 
To demonstrate the performance of various attribution models, we compare our DNAMTA model with three commonly used attribution models i.e. last touch attribution, Logistic Regression and HMM \cite{abhishek2012HMM} in our experiments:

\begin{itemize}
\item \textbf{LSTM} is the fundamental LSTM model without attention mechanism. 

\item \textbf{DNAMTA} is the first version of our deep attribution model with attention mechanism. After getting the outputs from each LSTM module, we will calculate the attention weights based on Formula(\ref{eq:att}), later we will use the re-weighted LSTM outputs as a path representation vector for binary classification modeling. 
\item \textbf{DNAMTA with time decay} is the second version of our deep attribution model. Besides incorporating attention mechanism, it also accounts for temporal-effect in attribution. The time decay weighted attention calculation formula is followed by Formula(\ref{eq:att_decay}). For simplification, we assume time decay parameter stays the same for all channels, but each channel (e.g. page search) could have its own time decay parameter. 
\item \textbf{Fusion DNAMTA} is the third version of our deep attribution model. Built on the top of time decayed DNAMTA, control variables such as user activity will be learned as a feature representation vector in another neural network. A fused representation vector is generated by concatenating touchpoint path vector and control variable vector, and it will be used for classification task based on the Formula(\ref{eq:fusion}). 
\end{itemize}

\par  We use TensorFlow 1.2.0 \cite{abadi2016tensorflow} and Python 3.0 for all deep model implementation, and sklearn 0.18.1, pomegranate for baseline model implementation.  All the comparison experiments are run on GPU Tesla K80 and CPU. For LSTM model we choose to use stochastic Adam gradient descent \cite{kingma2014adam} for training. In deep model, both the dimension of hidden size and attribution dimension (a.k.a. contextual vector $u$'s dimension) are 64. We use 3 hidden layers. During training process, a validation data set is hold out for hyper parameter tuning, and the model training process stops when the validation loss stops improving.



\subsection{Results}
\par In Table \ref{tab:prediction}, we report the prediction performance of all attribution models on testing dataset. We can observe that DNAMTA fusion model successfully utilizes both time and touchpoint dependent representation and confounding factors, and it achieves the highest prediction accuracy and AUC. Besides, on comparing with other models, we find that deep model with attention can generally improve the model prediction performance, which indicates the impact of attention mechanism in dynamic sequence path classification task, as attention can smartly reconsider touchpoint contextual dependencies and reallocate these touchpoint contributions. 
\par As we mentioned in previous section, model prediction is not the only goal for attribution modeling. From the perspective of representation learning, a good representation for dynamic path is good for future statistical inference and strategy decision making. The path representation vector from last touchpoint attribution model is simple without modeling, but it does not capture the time dependency between each touchpoint. If both a long touchpoint sequence and a short one ends with up the same touchpoint, these two paths will be considered same in the last touchpoint prediction model. For logistic regression, the path representation vector considers the touchpoint content information and time information, but the dimension of this vector can be dramatically high and sparse when our predefined observation time window grows. For our dataset that spans touchpoint data of 57 days, the feature dimension in logistic regression is 342. However, in DNAMTA model, the path representation dimension is only 64 and also achieves better prediction performance than logistic regression does, which shows us the efficiency of representation provided by DNAMTA. 
\par Similar to approach and arguments in \cite{mnih2014ram}, both the number of parameters in our model and the amount of computation it performs can be controlled independently of the size of the path if we fix the length of the customer path that is considered. Hence it is easily scalable with any size of data. In the case where we do not fix the path length, the computational demands scale linearly with the length of the path in consideration.


\subsection{Modified Attribution Score with Attention}
\par We propose a novel usage of the attention scores by incorporating it with traditional attribution score calculation \cite{kannan2013Experiment}: fractional attribution score and incremental attribution score. 
\vspace*{-.2cm}
\begin{itemize}
\item \textbf{Incremental score} We estimate the impact of a specific channel on the conversion by calculating the difference in conversion probabilities with and without the channel.

\item \textbf{Fractional score}  We normalize all incremental scores of each channel for each path, and aggregate all incremental contributions at channel level as the fractional score. 

\item \textbf{Attention based score} Attention values learned from deep model can be directly used as fractional score, as it serves as the contribution of each touchpoint after accounting for the interaction between each other.
\end{itemize}

\begin{table}[htb]
\centering
\caption{Fractional attribution values for different advertising channel}
\label{frac}
\small
\begin{tabular}{|p{1.1cm}|p{.5cm}|p{.5cm}|p{.5cm}|p{1cm}|p{1cm}|p{1cm}|}
\hline
           & LTA    & LR     & LSTM & DNAMTA & \multicolumn{1}{c|}{\begin{tabular}[c]{@{}c@{}}DNAMTA \\ timedecay\end{tabular}}& \multicolumn{1}{c|}{\begin{tabular}[c]{@{}c@{}}DNAMTA \\ fusion\end{tabular}}\\ \hline
Display    & 0.392 & 0.538 &  0.642    &  0.448 & 0.398       & 0.411     \\ \hline
Email      & 0.383 & 0.241 &   0.174   &  0.362  & 0.384         & 0.372      \\ \hline
PaidSearch & 0.225 & 0.221 &   0.184  &  0.190  & 0.218         & 0.217      \\ \hline
Total      &  1     & 1      &   1   &    1 & 1              & 1           \\ \hline
\end{tabular}
\end{table}

\begin{table}[htb]
\centering
\caption{Incremental attribution values for different advertising channel}
\label{margin}
\small
\begin{tabular}{|p{1.1cm}|p{.5cm}|p{.5cm}|p{.5cm}|p{1cm}|p{1cm}|p{1cm}|}
\hline
           & LTA    & LR     & LSTM & DNAMTA & \multicolumn{1}{c|}{\begin{tabular}[c]{@{}c@{}}DNAMTA \\ timedecay\end{tabular}}& \multicolumn{1}{c|}{\begin{tabular}[c]{@{}c@{}}DNAMTA \\ fusion\end{tabular}}\\ \hline
Display    &  0.325&   0.356  & 0.392 & 0.369 &   0.326& 0.341      \\ \hline
Email      &0.133 &    0.155  & 0.158  & 0.169 &  0.183& 0.180      \\ \hline
PaidSearch & 0.213&    0.162  & 0.131 & 0.176 &  0.206 &    0.207   \\ \hline
Total      & 0.671 &   0.673& 0.681 &   0.714& 0.715 &    0.728  \\ \hline
\end{tabular}
\end{table}

\begin{table}[htb]
\centering
\caption{Model prediction performance numerical values summary and comparison}
\label{tab:prediction}
\small
\begin{tabular}{|p{.9cm}|p{.5cm}|p{.5cm}|p{.5cm}|p{.5cm}|p{1cm}|p{.5cm}|p{.5cm}|}
\hline
           & LTA    & HMM & LR     & LSTM & DNAMTA & \multicolumn{1}{c|}{\begin{tabular}[c]{@{}c@{}}DNAMTA \\ timedecay\end{tabular}}& \multicolumn{1}{c|}{\begin{tabular}[c]{@{}c@{}}DNAMTA \\ fusion\end{tabular}}\\ \hline
Accuracy & 0.765 &0.766& 0.789 &   0.807   & 0.807    & 0.807         & 0.819     \\ \hline
AUC      & 0.800 &0.801& 0.846 & 0.841     &   0.855  & 0.851         & 0.879      \\ \hline
\end{tabular}
\end{table}
\vspace*{-.15cm}
\par In all attribution models, display accounts for the most contribution for customer conversion. Especially for logistic regression and DNAMTA model, both fractional and incremental attribution scores for Display are very high. While after incorporating time decay property, DNAMTA with timedecay and fusion model lowers the display attribution scores.  Indeed, customer has to be exposed to the product first before they can start their conversion journey. Display triggers the continuing advertising exposure, while display is usually less likely to show up closer to conversion. We didn't include HMM in the comparison table (\ref{margin}, \ref{frac}), because the attribution scores for HMM are quite similar to others. 
\par
\begin{figure}[htb]
\centering
\includegraphics[width=1.\linewidth]{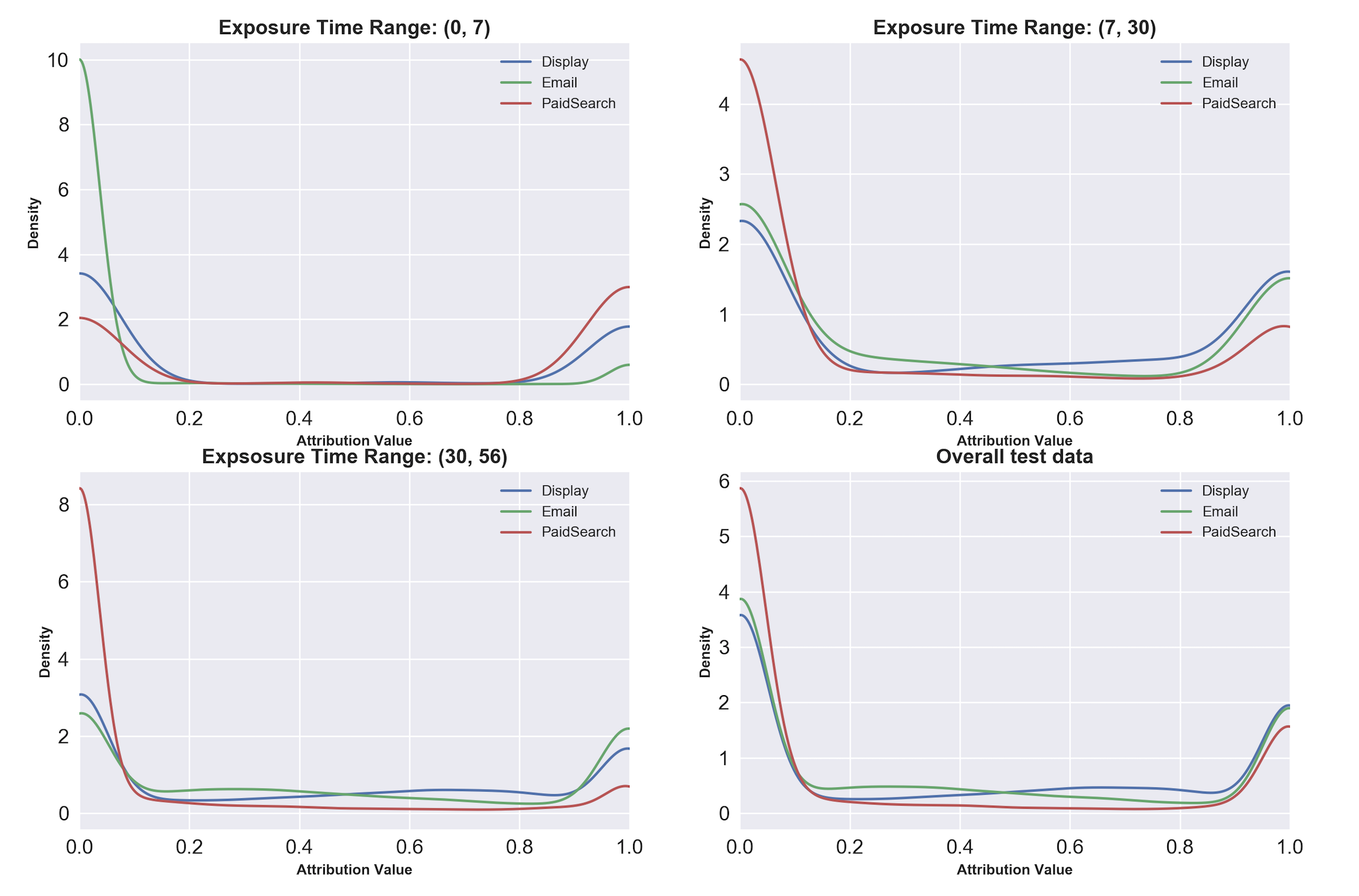}
\caption{Attribution estimate density distributions for each ad channel vary over different ad exposure time. The area under the curve of a density function represents the probability of getting specific attribution values between a range. The number of days until customer convert  ranges from top left to bottom right are: 0-7, 7-30, 30-56, 0-56. Paid search has relatively high impact within the first week,but this contribution decreases for long time exposure of ads.  }
\label{fig:time_dependency}
\end{figure}

 Figure(\ref{fig:time_dependency}) visualizes attribution density distributions for each touchpoint over various ad exposure lag. Overall, display accounts for the most conversion contribution, but among customers with different ads time exposure, touchpoint contribution distributions vary. For example, paid search has relatively high impact within the first week, but this contribution decreases for long time exposure of ads. As we mentioned in Section [\ref{notation}], DNAMTA is capable of capturing the underlying structure of touchpoints and their conversion contributions. 
\par
\begin{figure}[htb]
\centering
\subfigure[Display Click]{
\includegraphics[width=0.3\linewidth]{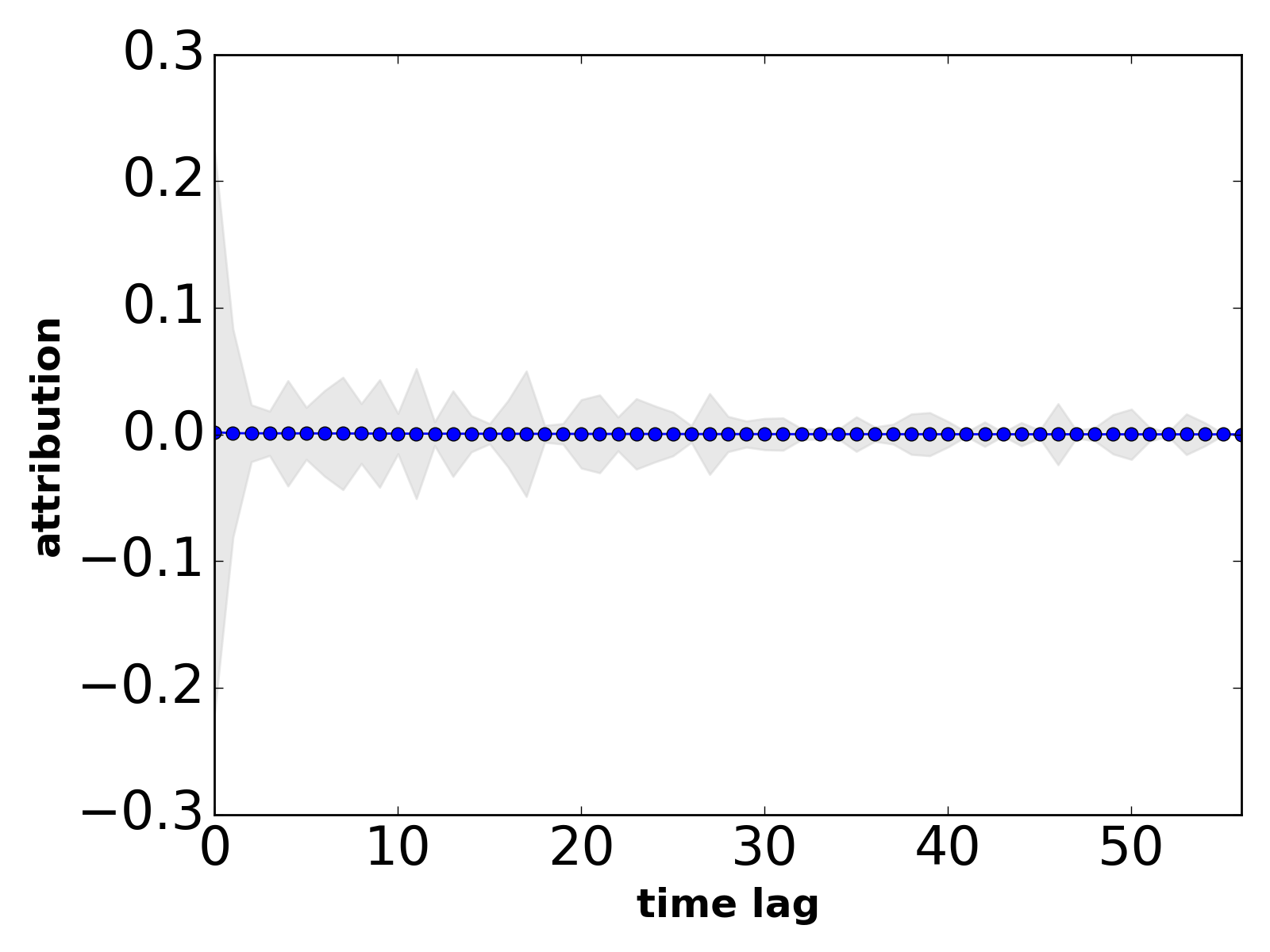}
\label{fig:DC}
}
\subfigure[Display Impression]{
\includegraphics[width=0.3\linewidth]{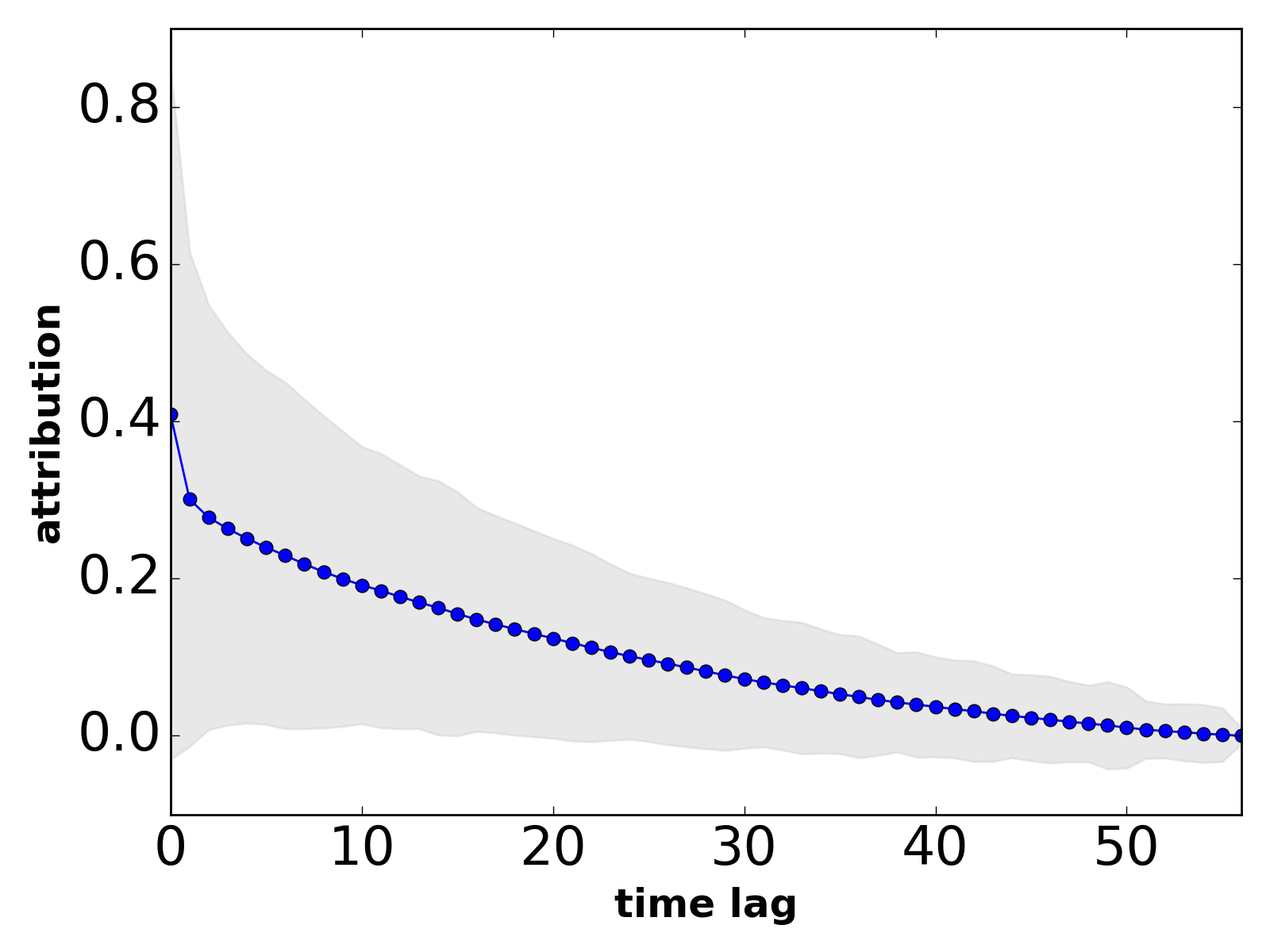}
\label{fig:DI}
}
\subfigure[Email Click]{
\includegraphics[width=0.3\linewidth]{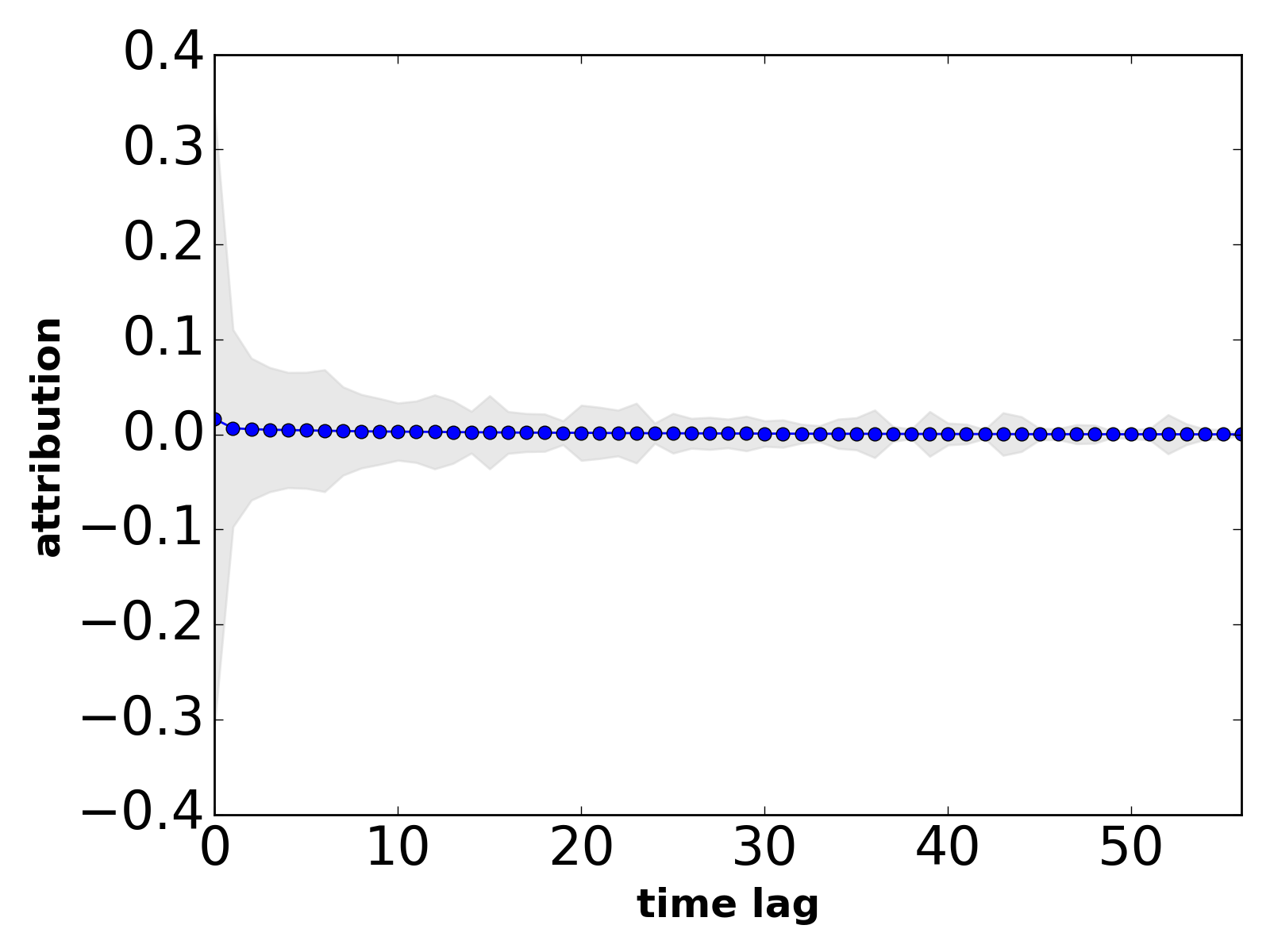}
\label{fig:EC}
}
\subfigure[Email open]{
\includegraphics[width=0.3\linewidth]{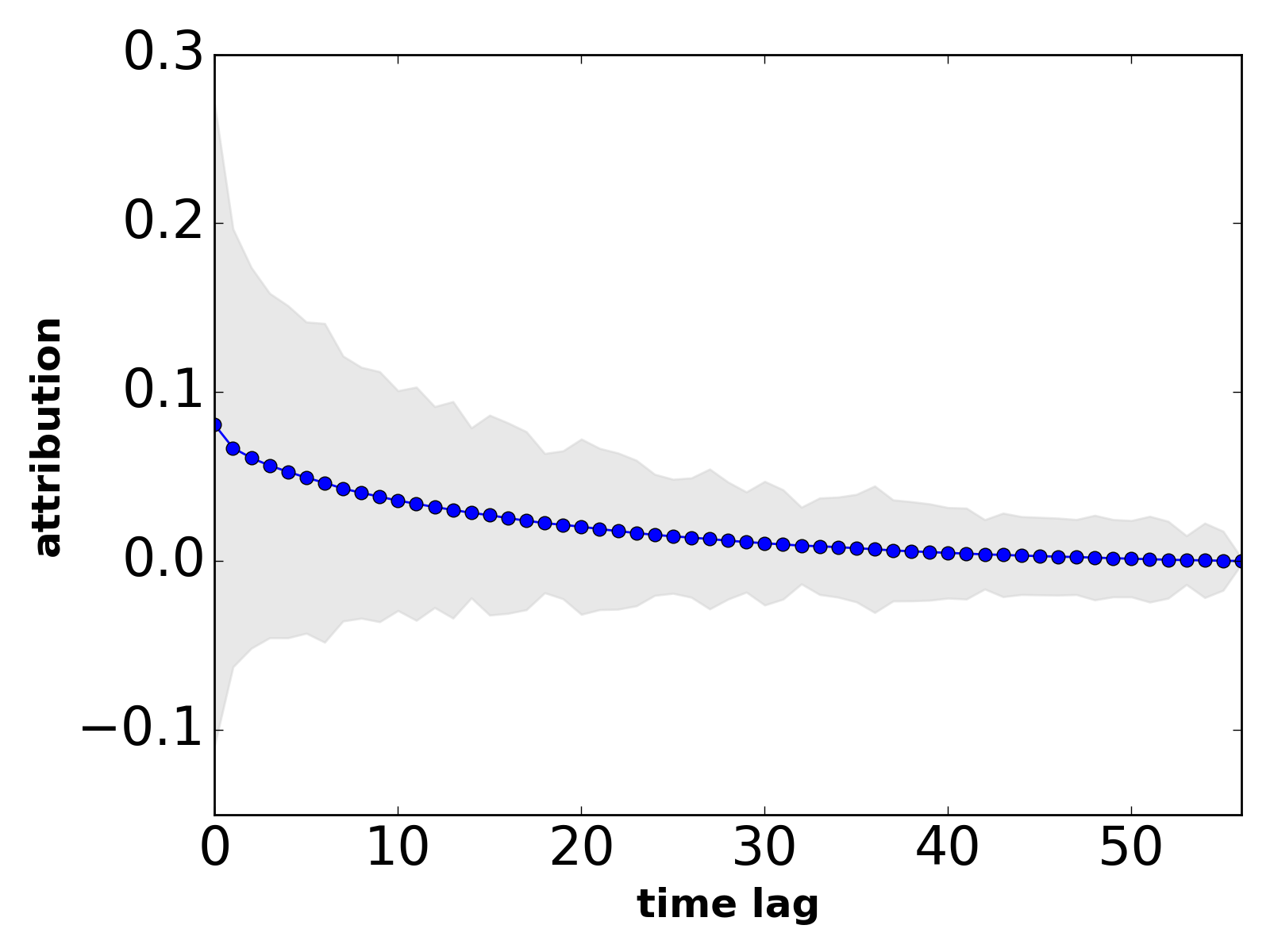}
\label{fig:EO}
}
\subfigure[Email Sent]{
\includegraphics[width=0.3\linewidth]{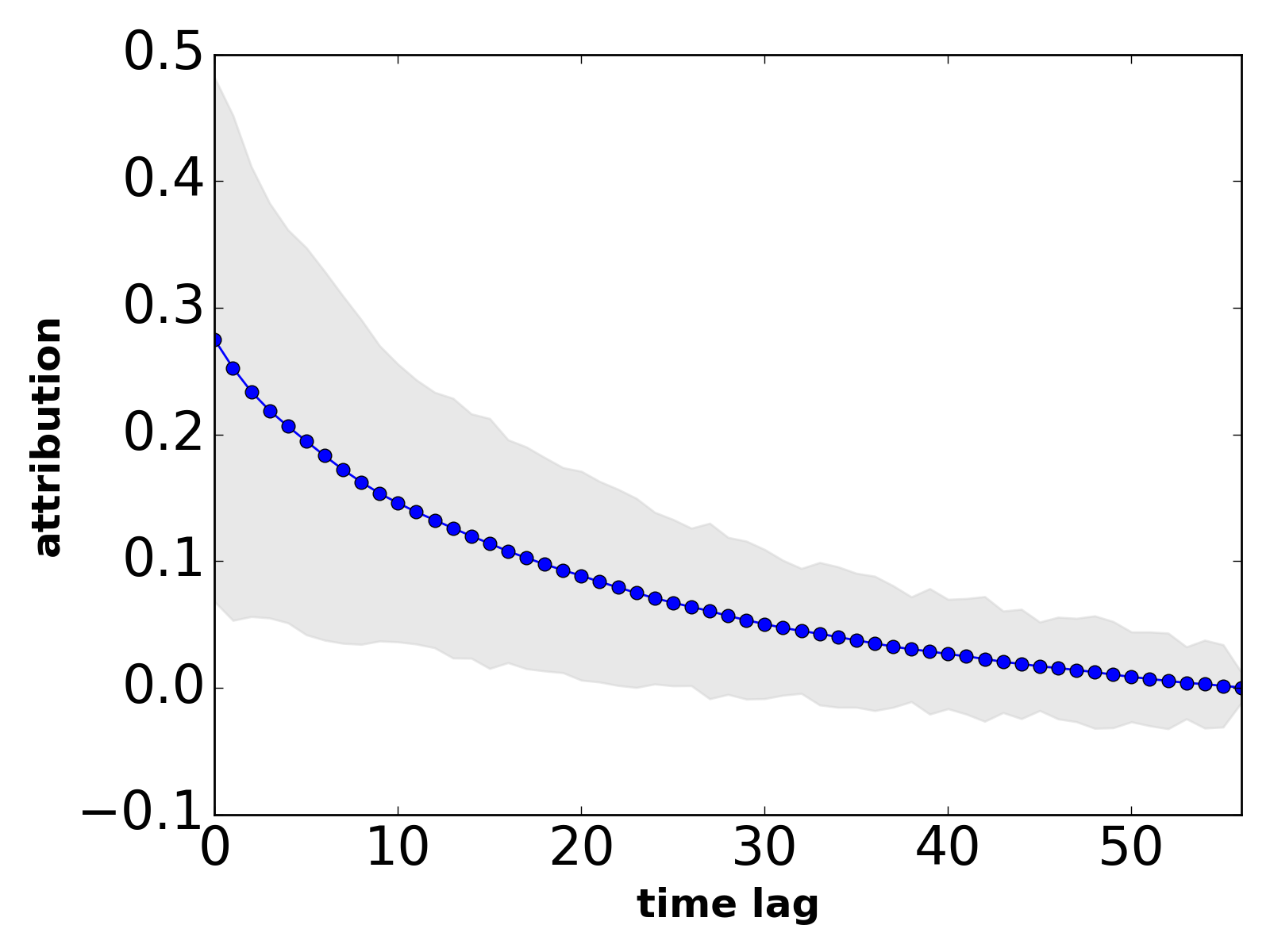}
\label{fig:ES}
}
\subfigure[Paid Search]{
\includegraphics[width=0.3\linewidth]{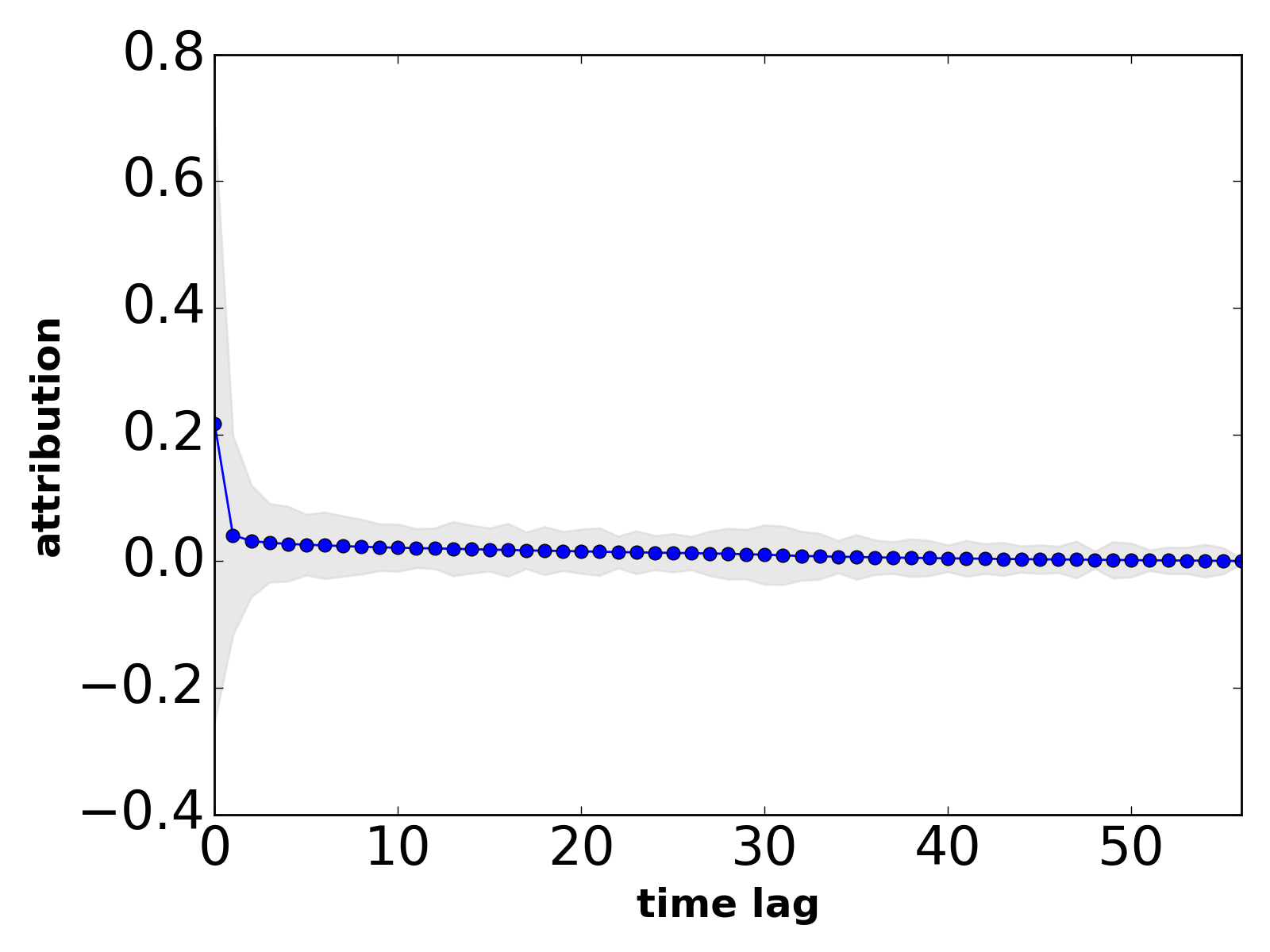}
\label{fig:PS}
}
\caption{Mean fractional attribution measured on y-axis decreases as time lag increases indicating the time decay property for all the touchpoints. Variance in mean fractional attribution indicated by the grey shadow area also has decreasing trend.}
\label{fig:time}
\end{figure}

 Figure(\ref{fig:time}) shows the time decay of attribution for each touchpoint. As the time lag (the difference between observation time and the end time of path) increases, the attribution for each touchpoint decreases. It confirms the time decay property for attribution score. The variance of on-average attribution score at specific time lag also has a decreasing trend as the time lag increases. The most latest advertising exposure may contribute a lot to customer's final conversion decision. 


\vspace{-.2cm}
\section{Conclusion and Future Work}
\par In this paper, we introduce DNAMTA, a deep neural network framework incorporating attention mechanism, by considering temporal effect and user characteristics through control variable adjustment. It aims to have a deeper understanding about the dynamic interactions between advertising channels and their contributions to customer conversion. For predictive task, DNAMTA surpasses some widely used attribution models as well as basic LSTM model. For interpretability, DNAMTA can also provide good insights of the relative touchpoint attribution estimates. 
\par Through the discussion in this paper, we also formalize attribution as a representation learning problem. Experiment results show that the dynamic path vector representation of dimension 64 from DNAMTA achieves better prediction performance compared to other attribution models. A good representation for dynamic advertising channels is not only good for prediction task and statistical inference, but also can be beneficial for transfer learning: transferring the domain knowledge and data-driven features to some other marketing problems with limited data observations. Marketers can thus allocate their budget spends on touchpoints in proportion to their contributions. 

\bibliographystyle{unsrt}
\bibliography{ref.bib} 
\end{document}